\documentclass{article}

\PassOptionsToPackage{numbers, compress}{natbib}
\usepackage[final]{nips_2017}

\usepackage[utf8]{inputenc} 
\usepackage[T1]{fontenc}    
\usepackage{hyperref}       
\usepackage{url}            
\usepackage{booktabs}       
\usepackage{amsfonts}       
\usepackage{nicefrac}       
\usepackage{microtype}      

\usepackage{dsfont}
\usepackage{subcaption}
\usepackage{graphicx}
\usepackage{amsmath}
\usepackage{amssymb}

\title{Roll-back Hamiltonian Monte Carlo}

\author{
  Kexin Yi \\
  Department of Physics\\
  Harvard University\\
  Cambridge, MA, 02138 \\
  \texttt{kyi@g.harvard.edu} \\
  \And
  Finale Doshi-Velez \\
  School of Engineering and Applied Sciences \\
  Harvard University \\
  Cambridge, MA, 02138 \\
  \texttt{finale@seas.harvard.edu}\\
}

\begin{document}

\maketitle

\begin{abstract}
We propose a new framework for Hamiltonian Monte Carlo (HMC) on truncated probability distributions with smooth underlying density functions. Traditional HMC requires computing the gradient of potential function associated with the target distribution, and therefore does not perform its full power on truncated distributions due to lack of continuity and differentiability. In our framework, we introduce a sharp sigmoid factor in the density function to approximate the probability drop at the truncation boundary. The target potential function is approximated by a new potential which smoothly extends to the entire sample space. HMC is then performed on the approximate potential. While our method is easy to implement and applies to a wide range of problems, it also achieves comparable computational efficiency on various sampling tasks compared to other baseline methods. RBHMC also gives rise to a new approach for Bayesian inference on constrained spaces.
\end{abstract}

\section{Introduction}

Markov Chain Monte Carlo (MCMC) provides an important family of sampling tools for Bayesian statistical inference \cite{Brooks2011}. Among various MCMC methods, Hamiltonian Monte Carlo (HMC) is one of the most widely adopted samplers for its ability to reduce random walk behavior and efficiently explore the sample space \cite{Neal2011, Hoffman2014, Betancourt2017, Carpenter2016}. Given a smooth target distribution $P(\mathbf{x})$, HMC maps it to a potential function $U$ by
\begin{equation}
U(\mathbf{x}) = - \log P(\mathbf{x})
\end{equation}
and proposes new samples according to the motion of a particle moving under this potential. The particle's dynamics is governed by the Hamiltonian equations of motion \cite{Goldstein1965}
\begin{equation}
\frac{d\mathbf{p}}{dt} =\nabla_{\mathbf{x}} H(\mathbf{p,x}) \qquad \frac{d\mathbf{x}}{dt} = -\nabla_{\mathbf{p}} H(\mathbf{p,x})
\label{eqn:Hamiltonian_equations_of_motion}
\end{equation}
where $\mathbf{x}$ is the spatial position of the particle in the sample space, $\mathbf{p}$ is the momentum, and $H$ is the Hamiltonian of the particle defined as $H(\mathbf{p, x}) = \mathbf{p}^2 / (2m) + U(\mathbf{x})$. At each iteration, a trajectory of the particle is simulated by discrete leapfrog updates with length $L$ and step size $\varepsilon$. A new sample is proposed at the end point of the trajectory (See \cite{Neal2011}).

In many applications, it is often desirable to draw samples from a truncated distribution other than the full distribution. For example, some Bayesian models, such as the Bayesian non-negative matrix factorization \cite{Schmidt2009}, assigns truncated prior to the latent variables to fix their values within a constrained region. Many common probability distributions by themselves are also defined on subregions of a larger space such as the exponential distribution and the Beta distribution. This kind truncated distributions can be characterized by an underlying density function $f$ and a region of interest (ROI), such that
\begin{equation}
P(\mathbf{x}) = f(\mathbf{x}) \mathds{1}_{\text{ROI}}(\mathbf{x}).
\end{equation}

Simulation of Hamiltonian dynamics requires computing $\nabla_\mathbf{x} U(\mathbf{x})$ at any point of the particle trajectory. Therefore, for truncated distributions, HMC is no longer effective due to the discontinuity at the truncation boundary, even though the density function might still be globally differentiable. When the particle steps out of ROI, the probability density $P$ drops to zero and $U$ boosts to infinity, causing the Hamiltonian dynamics to be ill-defined. The sampler might still be able to proceed by terminating the current trajectory and discarding the new sample, but this treatment will lead to lower acceptance rate especially in high dimensional spaces.

Several studies have been conducted on extending HMC to discontinuous distributions: Afshar \textit{et al.} \cite{Afshar2015} proposed the Reflective Hamiltonian Monte Carlo (RHMC), which incorporates reflection and refraction moves into the particle's dynamics at discontinuous boundaries. Pakman \textit{et al.} developed an HMC method that samples from piecewise Gaussian distribution by exactly solving the Hamiltonian equations of motion \cite{Pakman2013, Pakman2014}. Lan \textit{et al.} introduced spherical HMC for distributions on constraint boundary specified by q-norm of sample parameters \cite{Lan2014}. Other baseline methods for sampling from truncated distributions include rejection sampling \cite{MacKay1998} and Gibbs sampler \cite{Gelfand1992}.

The existing HMC-based methods either impose explicit boundary checks to the particle's dynamics \cite{Afshar2015} or require a specific form in the density function \cite{Pakman2014} or the domain boundary \cite{Lan2014}. Other non-HMC methods often suffer from random walk behavior \cite{Neal2011}. In this work, we introduce a new framework of HMC that is both easy to implement and applies to general distributions with arbitrary truncation boundaries. Motivated by the barrier function in constrained optimization \cite{Lueberger1973}, we introduce a sigmoid function to approximate the drop in probability density, which effectively builds an infinitely rising inclined surface in potential at the truncation boundaries. Meanwhile, the target distribution remains almost unchanged within the region of interest. We call this method Roll-back Hamiltonian Monte Carlo (RBHMC), because the particle always rolls back whenever it reaches the boundary and climbs up the rising surface.

\section{Sigmoid approximator for the truncation boundary}

In this section we introduce the key ingredient of our method, which is to use a sharp sigmoid function to approximate the probability drop at the truncation boundary. It is assumed that the underlying density function $f(\mathbf{x})$ is continuous and differentiable over the entire sample space, which we call smooth in this context. We begin with the one-dimensional unit step function, and then generalize the approximator to arbitrary boundary shapes in higher dimensions.

\subsection{Sigmoid approximation for the 1D step function}

On the real line, a truncated distribution can be characterized by the density function $f(x)$ plus a unit step function at the truncation point
\begin{equation}
P(x) = f(x) u(x)
\end{equation}
where $u(x) = 0$ when $x < 0$, $u(x) = 1$ when $x \geq 0$.
The ROI corresponding to the above expression is $(0, \infty)$. In general one can add multiple step functions at various locations to specify multiple truncation points. As an example, the exponential distribution can be written as $P(x) = \lambda \exp (-\lambda x) u(x)$. 

\begin{figure}
	\centering
	\begin{subfigure}{0.32\textwidth}
		\centering
		\includegraphics[width = \linewidth]{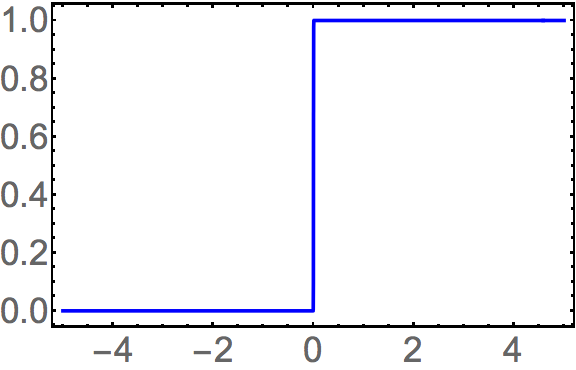}
		\caption{}
	\end{subfigure}
	\begin{subfigure}{0.335\textwidth}
		\centering
		\includegraphics[width = \linewidth]{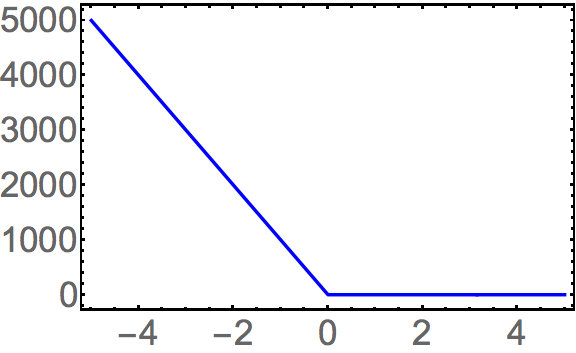}
		\caption{}
	\end{subfigure}
	\begin{subfigure}{0.32\textwidth}
		\centering
		\includegraphics[width = \linewidth]{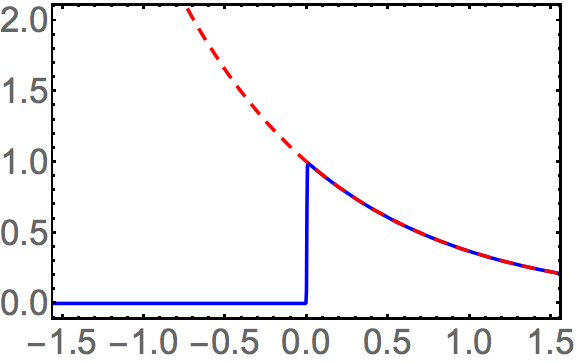}
		\caption{}
	\end{subfigure}
	\caption{(a) Sigmoid function with $\mu = 1000$. (b) Boundary potential $U_{\sigma}$ associated with the sigmoid factor. (c) Continuous approximation of exponential distribution (blue curve) by multiplying the sigmoid factor to the underlying density function (red dashed curve).}
	\label{fig:sig_approx}
\end{figure}

The sigmoid function 
\begin{equation}
\sigma(x; \mu) = \frac{1}{1 + \exp (-\mu x)}.
\end{equation}
is continuous and differentiable on the entire real line, whose value monotonically increases from 0 at $-\infty$ to 1 at $+\infty$. When $\mu$ goes to $+\infty$, the sigmoid function converges to the unit step function located at the origin (figure \ref{fig:sig_approx}a). This observation suggests that a smooth approximation of the truncated distribution can be obtained by replacing $u(x)$ with $\sigma(x; \mu)$ and choosing a reasonably large $\mu$. An example of the approximated exponential distribution is shown in figure \ref{fig:sig_approx}c. The sigmoid function is able to sharply flatten the exponentially rising head for negative $x$ (red dashed curve), but on the other hand, causes very little variations on the target density function when $x > 0$. The precision of approximation can always be matched to the requirement of any numerical purpose by increasing the value of $\mu$. In practice, one can invert the sign or add constant shift to $x$ in order to create approximated truncation at any other locations.

\subsection{Generalization to multi-dimensional sample spaces}

In the 1D example, the sudden jump in the sigmoid function $\sigma(x; \mu)$ occurs at $x = 0$. We can use this property to generalize the approximator to arbitrary region boundaries in higher dimensions by introducing an intermediate mapping $g(\mathbf{x})$ such that the ROI is given by
\begin{equation}
\text{ROI} = \{ x : g(\mathbf{x}) > 0 \}.
\label{eqn:def_g(x)}
\end{equation}
Similar to the unit step function in 1D, the following modified sigmoid function
\begin{equation}
\sigma(g(\mathbf{x}); \mu) = \frac{1}{1 + \exp(-\mu g(\mathbf{x}))}.
\end{equation}
continuously approximates the region's indicator function $\mathds{1}_{\text{ROI}}(\mathbf{x})$. For example, assume we want to approximate an arbitrary smooth density function $f(x, y)$ in $\mathbb{R}^2$ truncated by a unit circle centered at the origin. We can choose $g$ to be $g(x, y) = 1 - x^2 - y^2$. Then $f(x, y) g(x, y)$ gives the smooth approximator of the target distribution.

In principle, The choice of $g(\mathbf{x})$ is ambiguous as long as it specifies the correct ROI. However, in order to create a sharp jump in probability at the region boundary, we desire $g(\mathbf{x})$ to be sensitive to change in each component of $\mathbf{x}$ at the boundary. We will discuss the numerical requirements in the following section.

\section{Roll-back Hamiltonian Monte Carlo}
With the sigmoid trick, we can now simulate the Hamiltonian dynamics and apply HMC to draw samples from the smooth distribution. In this section, we discuss the particle dynamics under the new approximate potential, and address some numerical issues associated with the simulation of Hamiltonian dynamics.

\subsection{Hamiltonian dynamics near ROI boundary}

The potential function associated with the approximate distribution is given by
\begin{equation}
\tilde{U}(\mathbf{x}) = -\log f(\mathbf{x}) + \log [ 1 + \exp(-\mu g(\mathbf{x})) ]
\end{equation}
where the second term is called the boundary potential and denoted as $U_{\sigma}$ because it originates from the sigmoid approximator for the truncation boundary. The term has the following desired properties. First, its value remains close to zero when $x \in \text{ROI}$. This means the particle dynamics within the ROI is not affected and HMC updates preserve the original target distribution. Second, the potential puts up an infinitely rising inclined surface at the boundary (figure \ref{fig:sig_approx}b). When the particle steps out of ROI, it first climbs up the inclined surface and then quickly rolls back in a reflective fashion. Without having to restart the current trajectory, the particle's motion is restricted within the region of interest.

\begin{figure}
	\centering
	\begin{subfigure}{0.49\textwidth}
		\centering
		\includegraphics[width = \linewidth]{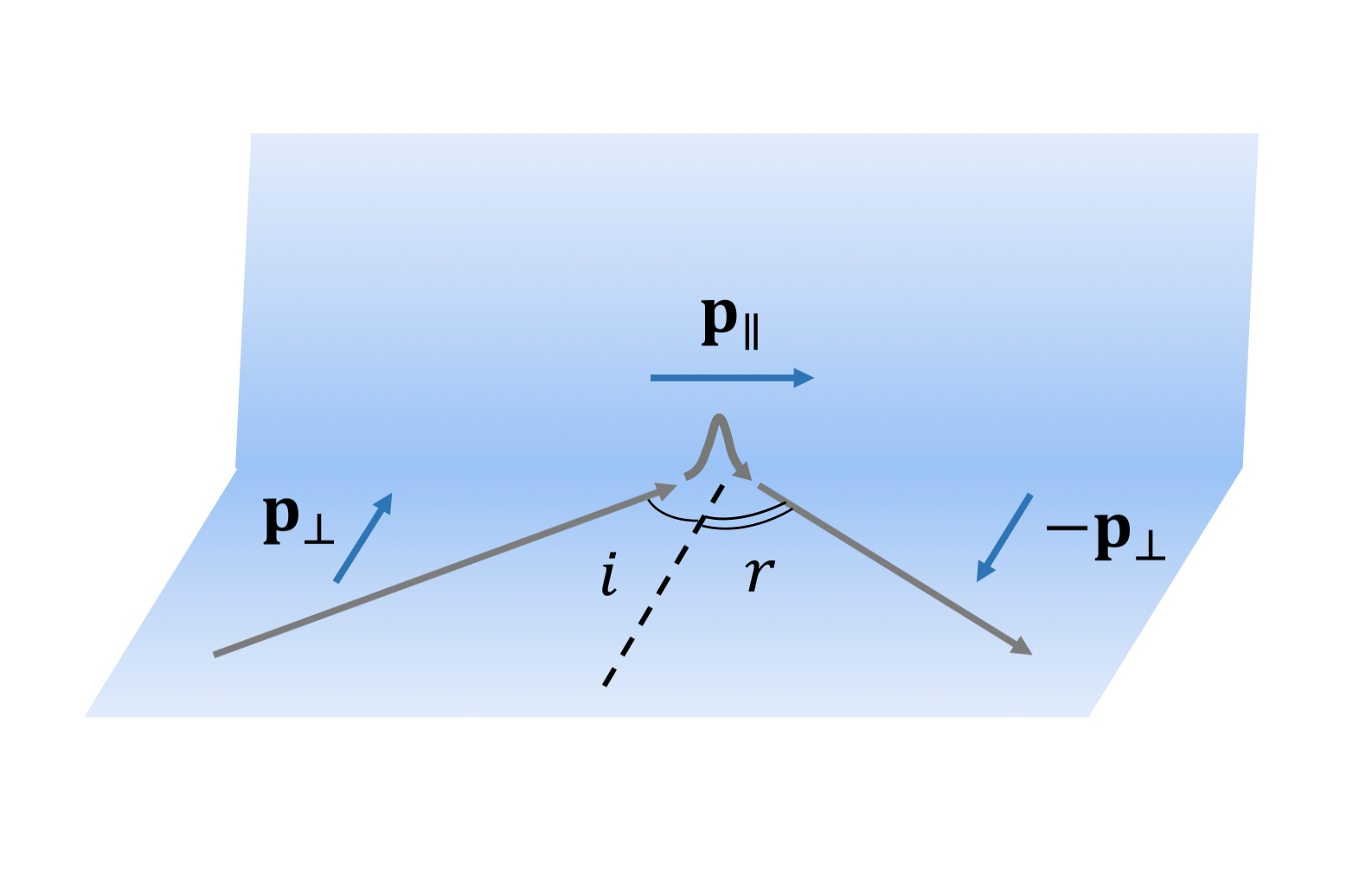}
		\caption{}
	\end{subfigure}
	\begin{subfigure}{0.49\textwidth}
		\centering
		\includegraphics[width = \linewidth]{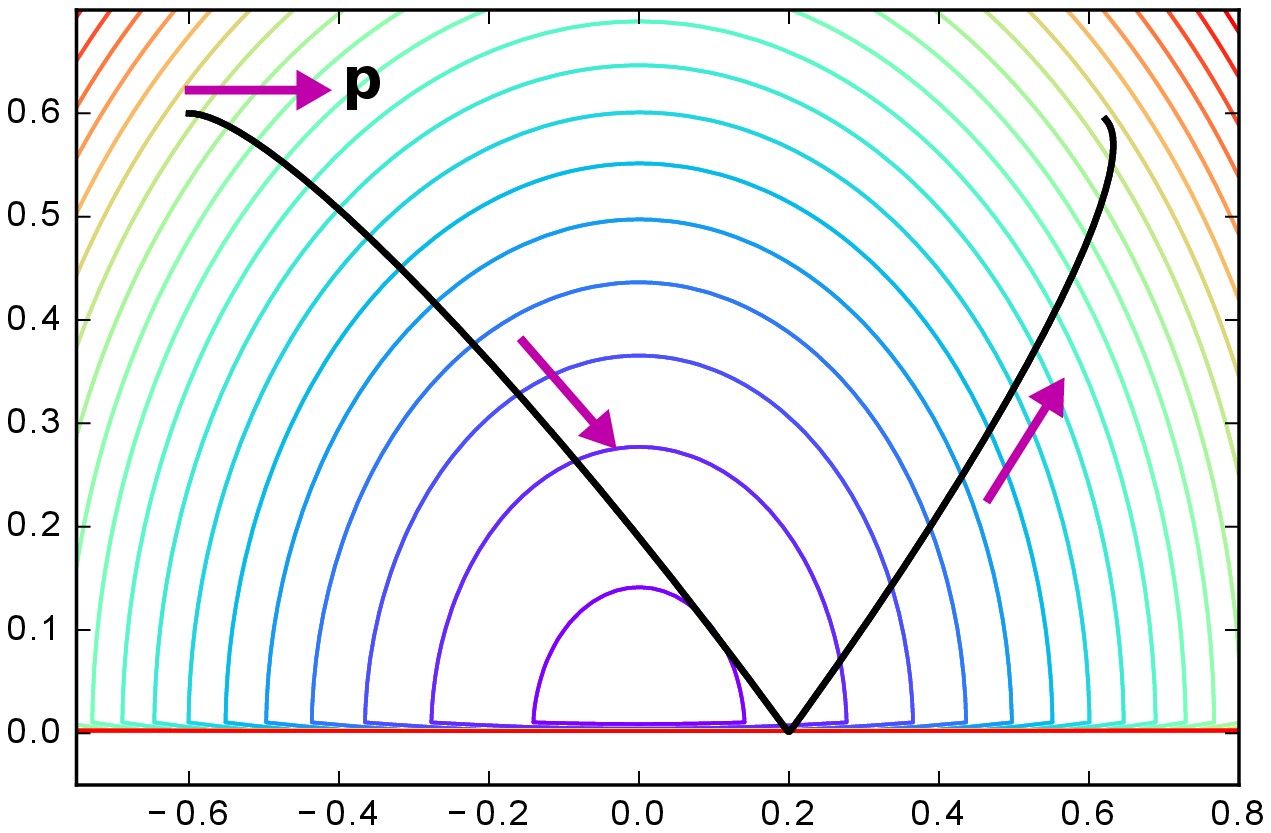}
		\caption{}
	\end{subfigure}	
	\caption{(a) Schematic sketch of roll-back dynamics at ROI boundary. Under the desired approximation setup, the incidence angle $i$ equals to the reflection angle $r$. (b) Simulated reflective trajectory on a 2D Gaussian distribution truncated by the $x$-axis. $\tilde{U}(x, y) = (x^2 + y^2) / 2 + \log[1 + \exp(-\mu y)], \mu = 1000$, step size $ \varepsilon = 0.002$.}
	\label{fig:Roll-back}
\end{figure}

We now consider the particle's dynamics near the truncation boundary. During each leapfrog step, the change in momentum caused by the boundary potential follows the direction of $-\nabla_{\mathbf{x}} \tilde{U}_{\sigma}(\mathbf{x})$, which always points to the perpendicular direction. When $\mu$ is very large, this term dominates the total gradient of $\tilde{U}$ outside the boundary, therefore the parallel component of momentum $\mathbf{p}_{\parallel}$ remains unchanged. On the other hand, conservation of energy requires the particle to have the same kinetic energy $\mathbf{p}^2/(2m)$ before and after rolling back, so the perpendicular component of momentum $\mathbf{p}_{\perp}$ must be inverted with the same magnitude. After all, the particle effectively experiences a reflection at the boundary (figure \ref{fig:Roll-back}).
\begin{equation}
\mathbf{p}_{\parallel} \rightarrow \mathbf{p}_{\parallel} \qquad \mathbf{p}_{\perp} \rightarrow -\mathbf{p}_{\perp}
\end{equation}

We note that RBHMC leads to the same reflective dynamics at truncation boundaries as RHMC \cite{Afshar2015}. However, the difference between the two methods lies in the fact that RHMC manually imposes the reflective move to the particle's dynamics, while in RBHMC it naturally emerges from the Hamiltonian dynamics under the approximate potential. 

\subsection{Numerical requirements on model and sampler parameters}

In practice, we don't have access to the exact continuous dynamics of the particle, and instead, the motion of the particle is simulated by the leapfrog algorithm based on discrete time steps. This leads to two major numerical issues that should be addressed for RBHMC to be correctly applied.

First, the slope of the inclined surface introduced by the boundary term should be large enough to dominate over the original potential. Consider the gradient of $\tilde{U}(\mathbf{x})$ which is proportional to the change of momentum during a leapfrog update at point $\mathbf{x}$:
\begin{equation}
\nabla \tilde{U}(\mathbf{x}) = - \frac{1}{f(\mathbf{x})} \nabla f(\mathbf{x}) - \frac{\mu}{1 + \exp[\mu g(\mathbf{x})]} \nabla g(\mathbf{x}).
\end{equation}
When $g(\mathbf{x}) < 0$ and $\mu$ is large, the second term (gradient of the boundary term) can be approximated by $- \mu \nabla g(\mathbf{x})$, which suggests that the slope of the inclined surface is proportional to $\mu$. To enforce the reflective roll-back and prevent the particle from going too far into the forbidden region, the second term should dominate over the first term at the boundary:
\begin{equation}
|\mu \nabla g(\mathbf{x})| \gg \left| \frac{1}{f(\mathbf{x})} \nabla f(\mathbf{x}) \right|.
\end{equation}
This condition suggests that other than $\mu$ being large, $g(\mathbf{x})$ should also have large or at least non-zero gradient at the boundary. Since the samples are drawn from the approximate distribution rather than the true distribution, a large $\mu$ is also required for the sake of a precise approximation and good sample quality.

Second, in order to simulate the full roll-back process, the leapfrog step size $\varepsilon$ can not be too large such that the particle should be able to feel the smooth rise in potential at the boundary. Assume the kinetic energy is given by $K(\mathbf{p}) = \mathbf{p}^2 / (2m)$, the step size should satisfy
\begin{equation}
\varepsilon \lesssim \frac{\sqrt{m}}{|\nabla g (\mathbf{x}) \mu|}.
\label{equ:eps_constraint}
\end{equation}
This upper bound can be derived from the following physical picture. By pushing forward one leapfrog step against the boundary, the particle should not be able to gain more potential energy than its incoming kinetic energy. Then $ \varepsilon |\nabla g (\mathbf{x}) \mu| |\mathbf{p}| / m \lesssim \mathbf{p}^2 / (2m)$, from which the above expression of the upper bound can be derived. Failing to meet this constraint will result in the particle picking up more perpendicular momentum after bouncing off the boundary, and lead to the wrong dynamics.

\subsection{Implementation}

We summarize the steps for implementing RBHMC as following. Given the density function of the target distribution $f(\mathbf{x})$, write down the potential function and its gradient $U(\mathbf{x}) = - \log f(\mathbf{x}), \nabla U(\mathbf{x}) = - \nabla f(\mathbf{x}) / f(\mathbf{x})$. For each truncation boundary, select $g(\mathbf{x})$ that satisfies equation \ref{eqn:def_g(x)} and $\nabla g(\mathbf{x}) \neq 0$ at every point of the boundary. Then choose a reasonably large $\mu$, add $\log[1 + \exp(- \mu \mathbf{x})]$ to the potential function and $- \mu \nabla g(\mathbf{x}) / [1 + \exp(\mu \mathbf{x})]$ to the gradient. Finally, the HMC step size $\varepsilon$ is chosen to satisfy equation \ref{equ:eps_constraint}, and the trajectory length $L$ is tuned to optimize the sampler's exploration horizon over the region of interest.

In contrast to our method, RHMC requires computing the first intersection between the ongoing trajectory and the truncation boundary at every leapfrog step \cite{Afshar2015}. This can be very hard to implement because it requires the analytic expression of the location of intersection starting from any $\mathbf{x}$ and $\mathbf{p}$, especially when there are multiple boundaries. Our method is free from this complication and can be easily applied to arbitrarily complex truncation setups, since multiple truncation boundaries can be combined by simply summing up the corresponding boundary potential terms.

\section{Empirical Study}

In this section, we post experiment results of RBHMC on various sampling tasks. We first show its performance as a general sampling method and demonstrate the ease of implementation on a 2D Gaussian distribution under various truncation boundaries. We also compare its computational efficiency with other HMC-based methods and investigate its application to Bayesian inference with an example of Bayesian non-negative matrix factorization.

\subsection{Sample from 2D truncated Gaussian distribution}

We first test RBHMC as a basic sampling tool on 2D truncated Gaussian distribution. In general, HMC's advantage is most obvious in higher dimensions, but here we stay in 2D in order to provide straight forward visualization of the sampling result with respect to the density function. 

\begin{figure}[t]
	\centering
	\begin{subfigure}{0.32\textwidth}
		\centering
		\includegraphics[width = \linewidth]{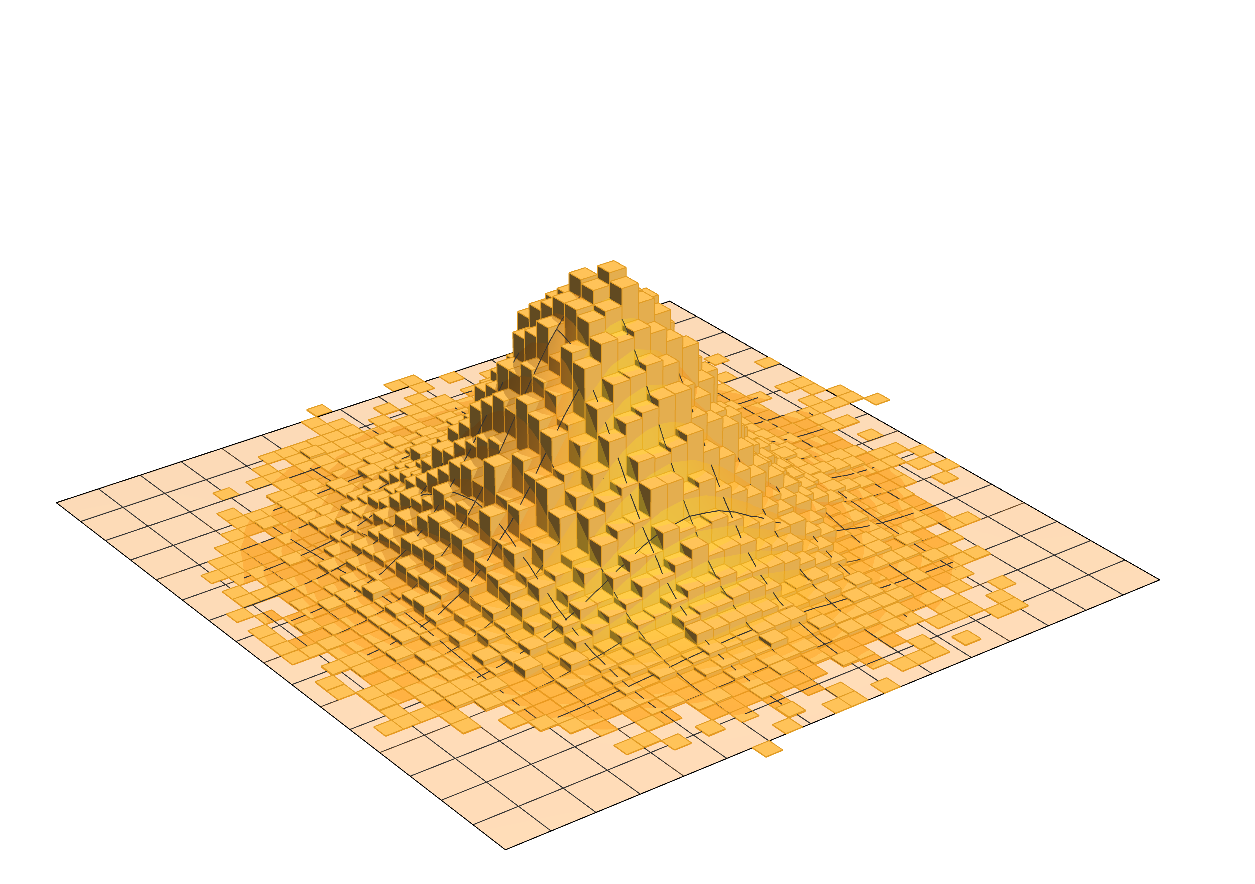}
		\caption{}
	\end{subfigure}
	\begin{subfigure}{0.32\textwidth}
		\centering
		\includegraphics[width = \linewidth]{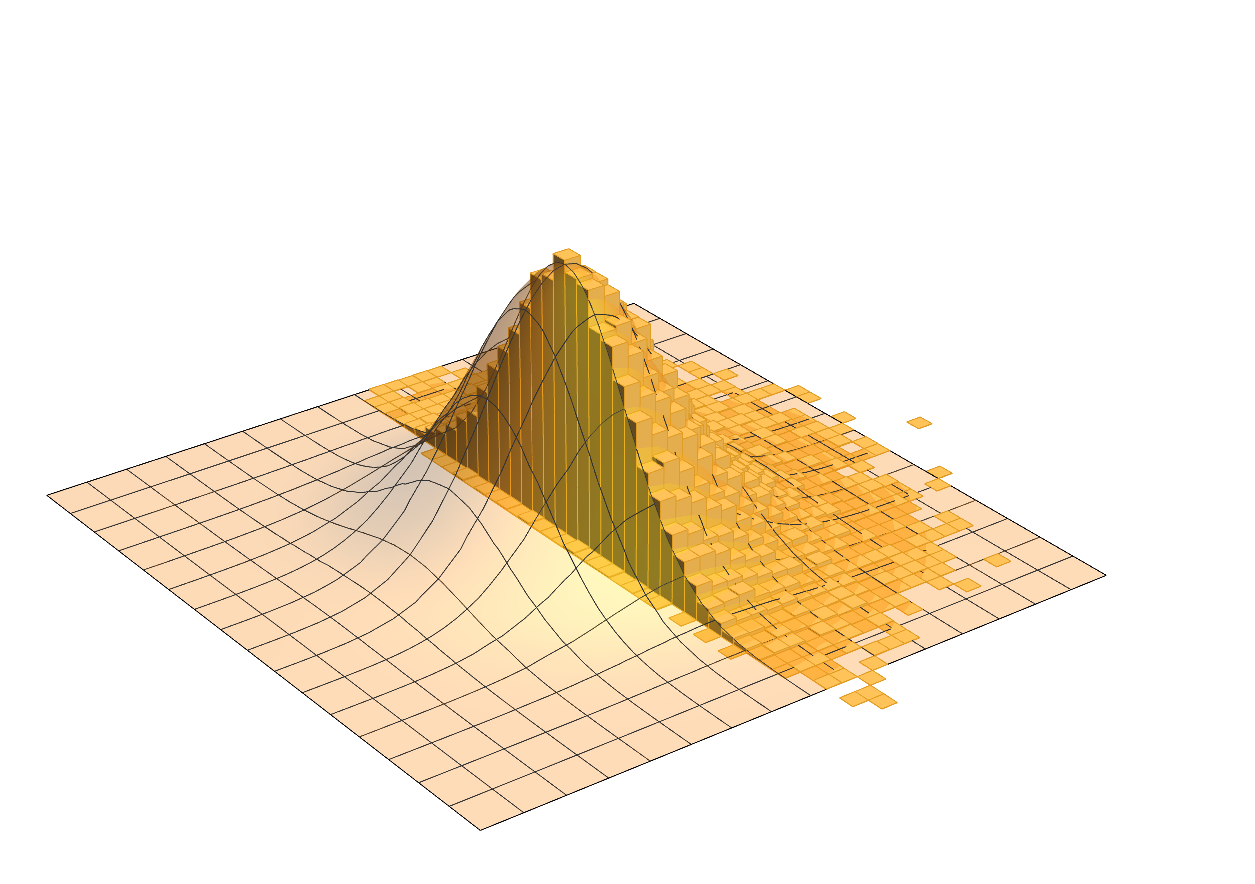}
		\caption{}
	\end{subfigure}	
	\begin{subfigure}{0.32\textwidth}
		\centering
		\includegraphics[width = \linewidth]{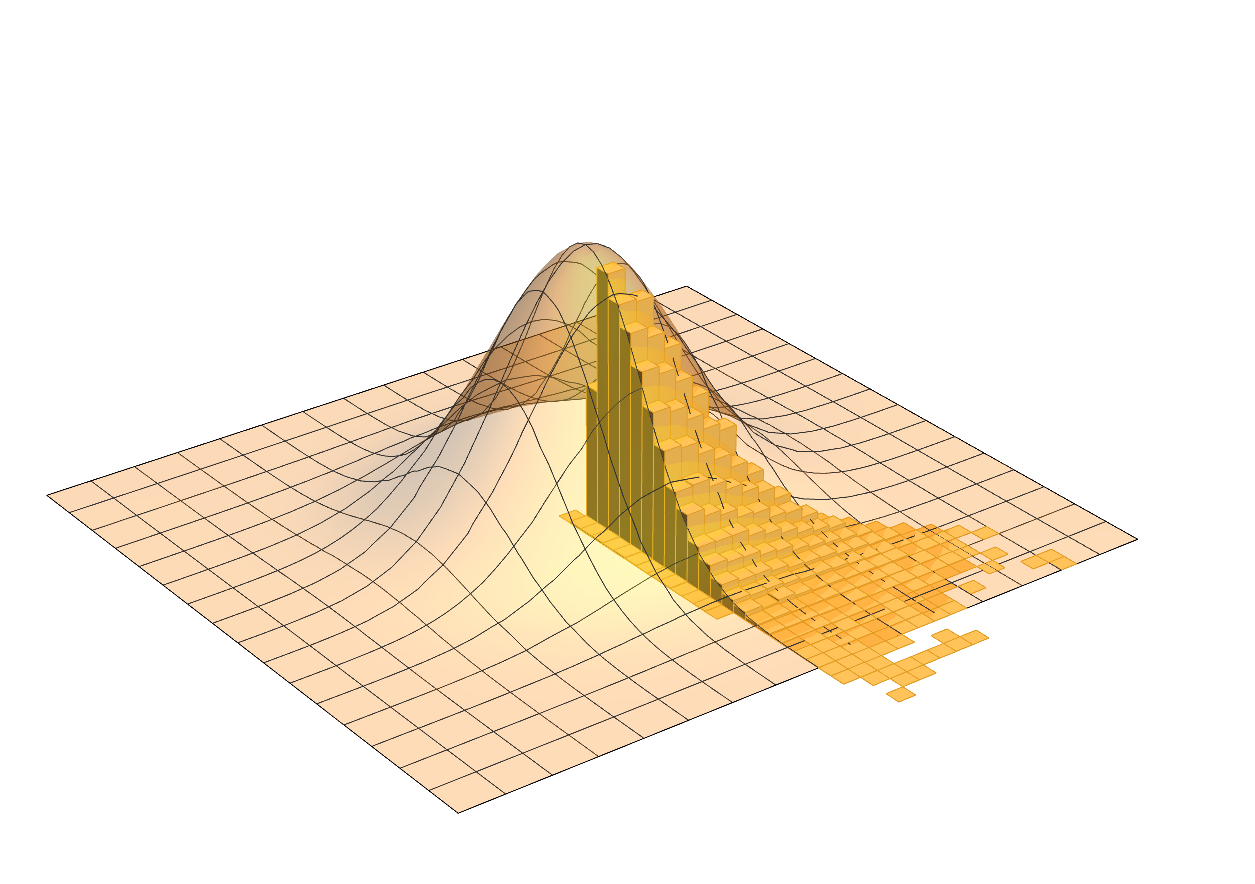}
		\caption{}
	\end{subfigure}	
	\begin{subfigure}{0.32\textwidth}
		\centering
		\includegraphics[width = \linewidth]{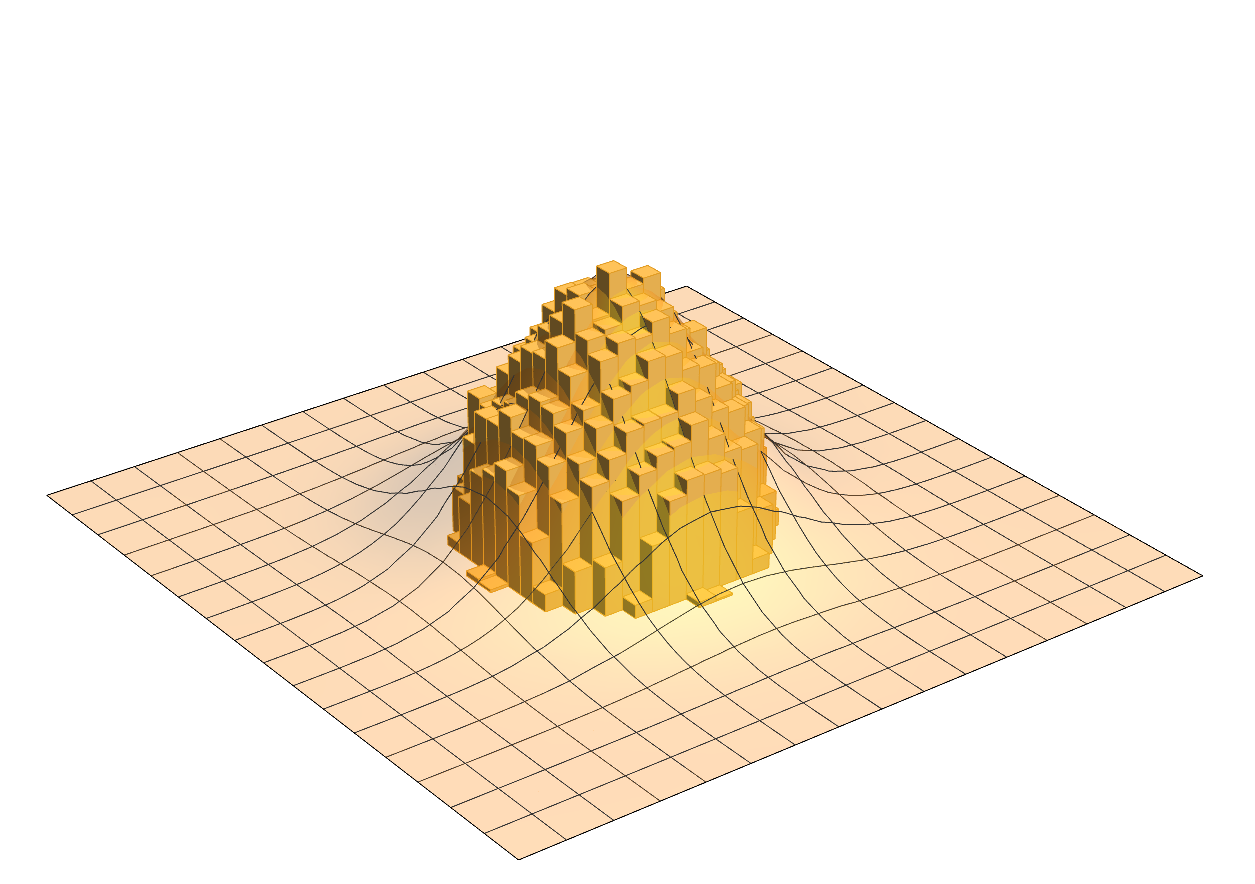}
		\caption{}
	\end{subfigure}	
	\begin{subfigure}{0.32\textwidth}
		\centering
		\includegraphics[width = \linewidth]{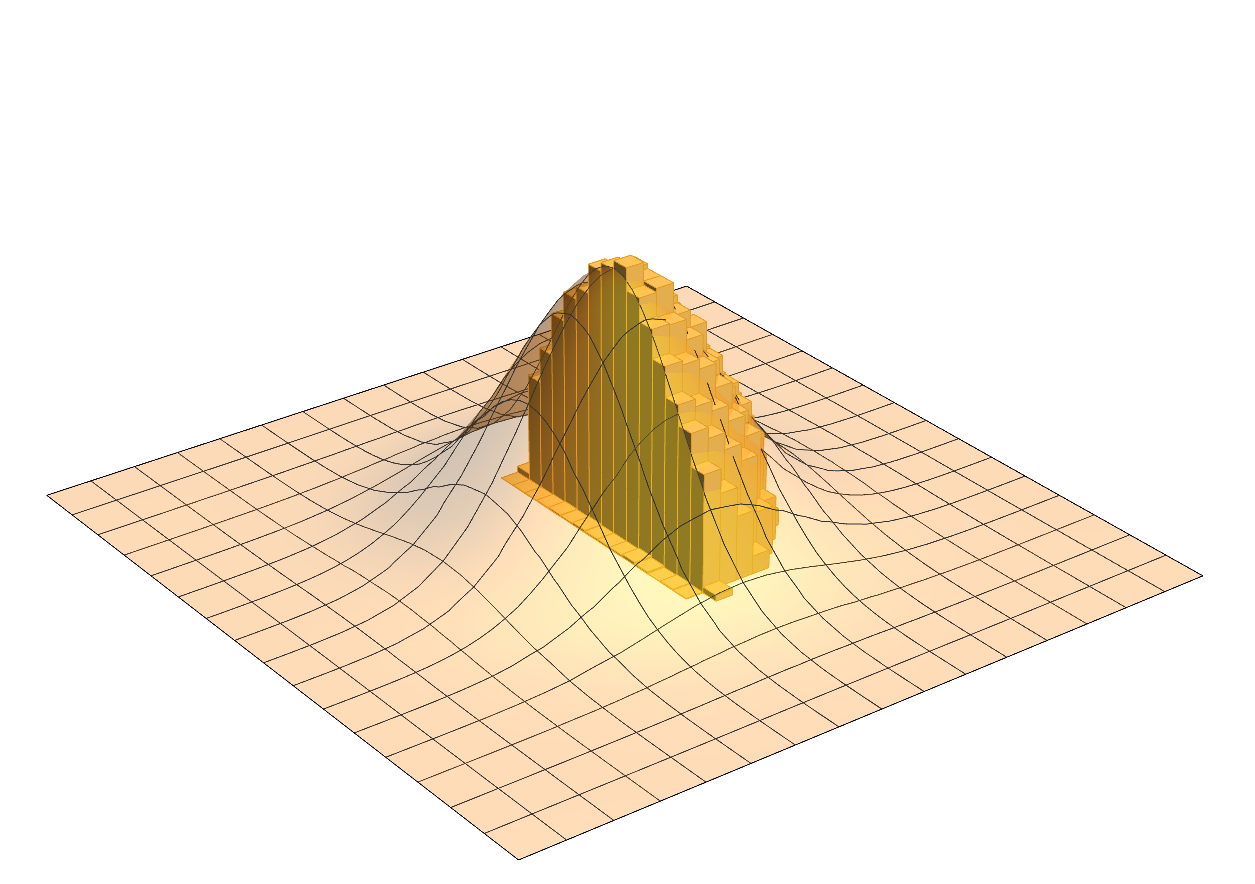}
		\caption{}
	\end{subfigure}	
	\begin{subfigure}{0.32\textwidth}
		\centering
		\includegraphics[width = \linewidth]{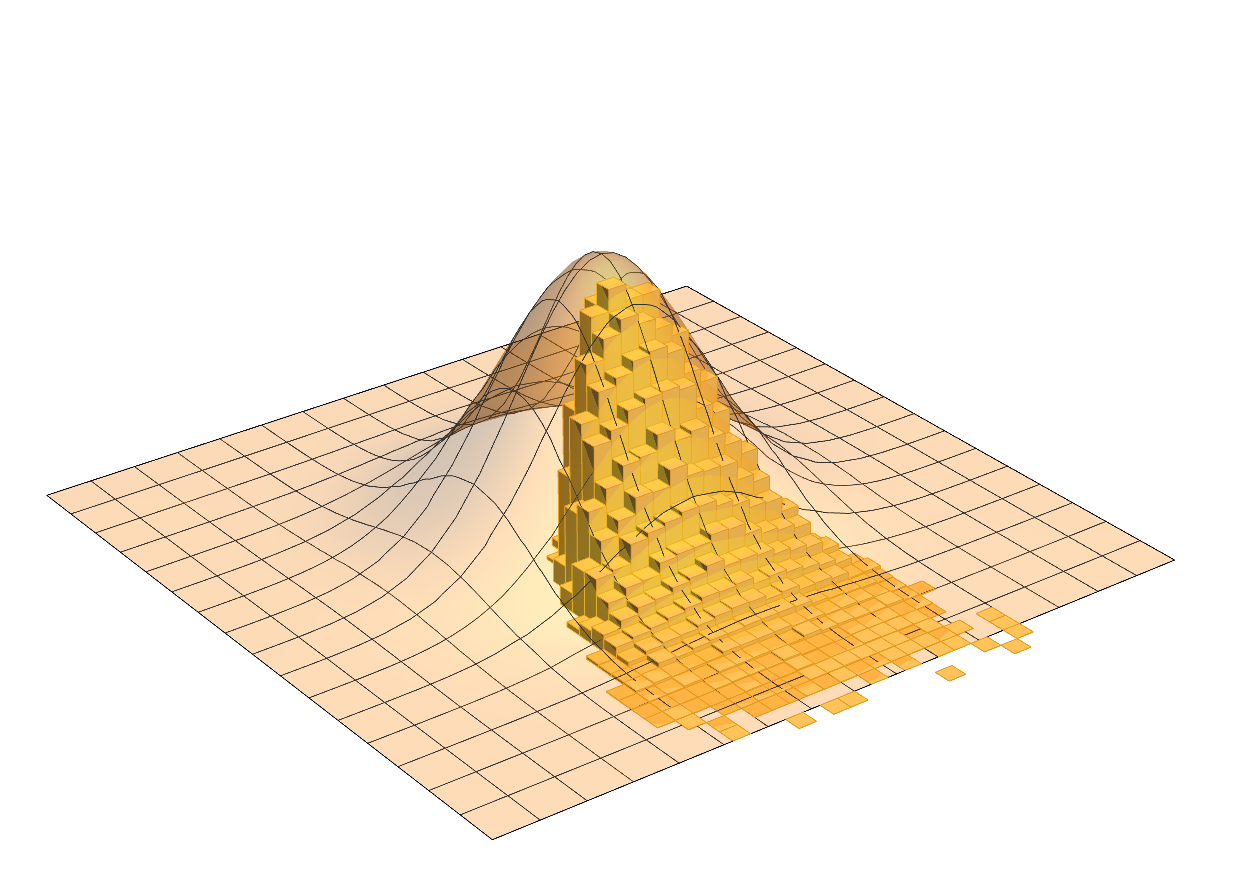}
		\caption{}
	\end{subfigure}	
	\caption{Sample histograms of 2D Gaussian distribution under various truncation boundaries drawn by RBHMC. Each square on the surface of density function corresponds to a $0.5 \times 0.5$ horizontal area. The expressions of the boundary terms associated with each subfigure and their gradients are listed in table \ref{tbl:boundary_terms}.}
	\label{fig:2dGauss}
\end{figure}

We run the sampler under 5 different truncation boundaries. As shown in figure \ref{fig:2dGauss}, all sample histograms are properly normalized and compared to the true density function, which is chosen to be the 2D standard Gaussian distribution $\mathcal{N}(\mathbf{0}, I)$. Figure \ref{fig:2dGauss}a shows the samples from the full distribution with no truncation. Figure \ref{fig:2dGauss}b, \ref{fig:2dGauss}c show the results for single and multiple linear truncation boundaries. Figure \ref{fig:2dGauss}d, \ref{fig:2dGauss}e show samples drawn from a disk and a half-disk centered at the origin. Figure \ref{fig:2dGauss}f shows result under an open parabolic truncation boundary. In all 6 cases, the sampler parameters are set to be $N = 100000, \mu = 500, \L = 100, \varepsilon = 0.004$, and the histograms all match to the density function very well. The boundary potential terms and their closed form gradients for each truncation boundary are listed in table \ref{tbl:boundary_terms}. As mentioned in the previous section, combining multiple truncation boundaries in RBHMC is a trivial task of adding up the boundary terms and their gradients. As a demonstration of this nice property, the semi-circle boundary in figure \ref{fig:2dGauss}e can be constructed by simply adding the boundary term for the straight line in \ref{fig:2dGauss}b to the boundary term for the full-circle in \ref{fig:2dGauss}d.  

\begin{table}
  \caption{List of $U_{\sigma}$ and $\nabla U_{\sigma}$ for various truncation boundaries}
  \label{tbl:boundary_terms}
  \centering
  \begin{tabular}{cccc}
    \toprule
    Subfigure  & $U_{\sigma}(x, y)$     & $\nabla U_{\sigma}(x, y)$ \\
    \midrule
    (a) & 0  & \textbf{0}     \\
    & & \\
    (b) & $\log[1 + \exp(- \mu y)]$ & $-\frac{\mu}{1 + \exp(\mu y)} \mathbf{e_y}$      \\
    & & \\
    (c) & $\log[1 + \exp(- \mu y)]$ & $-\frac{\mu}{1 + \exp(\mu (x - y))} \mathbf{e_x}$ \\
      & $\quad + \log[1 + \exp(- \mu (x - y))]$       & $\quad + \left[ \frac{\mu}{1 + \exp(\mu (x - y))} - \frac{\mu}{1 + \exp(\mu y)}\right]\mathbf{e_y} $  \\
    & & \\
    (d) & $\log[1 + \exp(- \mu (2 - x^2 - y^2))]$       & $\frac{2 \mu (x \mathbf{e_x} + y \mathbf{e_y}) }{1 + \exp(\mu (2 - x^2 - y^2))}$  \\
    & & \\
    (e) & $\log[1 + \exp(- \mu (2 - x^2 - y^2))]$       & $\frac{2 \mu x}{1 + \exp(\mu (2 - x^2 - y^2))} \mathbf{e_x}$  \\
      & $ \quad +\log[1 + \exp(- \mu y)]$ & $ \quad  +\left[ \frac{2 \mu y}{1 + \exp(\mu (2 - x^2 - y^2))} -\frac{\mu}{1 + \exp(\mu y)} \right] \mathbf{e_y}$ \\
    & \\
    (f) & $\log[1 + \exp(- \mu (x - y^2))]$       & $-\frac{\mu}{1 + \exp(\mu (x - y^2))} \mathbf{e_x}$  \\
      & & $\quad + \frac{2 \mu y}{1 + \exp(\mu (x - y^2))} \mathbf{e_y}$ \\
    \bottomrule
  \end{tabular}
\end{table}

\subsection{Evaluate computational performance}

We now test the computational performance of RBHMC using a similar experimental setup as in the RHMC paper \cite{Afshar2015}. The potential function is given by the following piecewise expression
\begin{align}
\begin{split}
U(\mathbf{x}) = \left\{
\begin{array}{ll}
\sqrt{\mathbf{x}^T\mathbf{A}\mathbf{x}} & \qquad |\mathbf{x}| \leq 3 \\
\infty & \qquad |\mathbf{x}| > 3
\end{array}
\right.
\end{split}
\end{align}
where $|\mathbf{x}|$ is the Euclidean norm of vector $\mathbf{x}$ and $\mathbf{A}$ is a positive definite matrix. We compare our method to RHMC and the baseline HMC in which current trajectory is terminated and new sample is rejected whenever the sampler steps out of ROI. Since the potential is centered at the origin, a Markov chain that has reached stationary distribution should have zero sample mean on all directions. For the HMC samplers, we take the mean of all drawn samples and compute the deviation from the origin to measure how well the sampler mixes. More specifically, we introduce the worst mean absolute error (WMAE) 
\begin{equation}
\text{WMAE} (\mathbf{x}^{(1)}, \mathbf{x}^{(2)}, ... \mathbf{x}^{(k)}) = \max\limits_{d = 1:D} \left| \frac{1}{k} \sum\limits_{s = 1}^k \mathbf{x}_d^{(s)} \right|.
\end{equation}
The same error measure is used in the experiments of \cite{Afshar2015}.

\begin{figure}
	\centering
	\begin{subfigure}{0.32\textwidth}
		\centering
		\includegraphics[width = \linewidth]{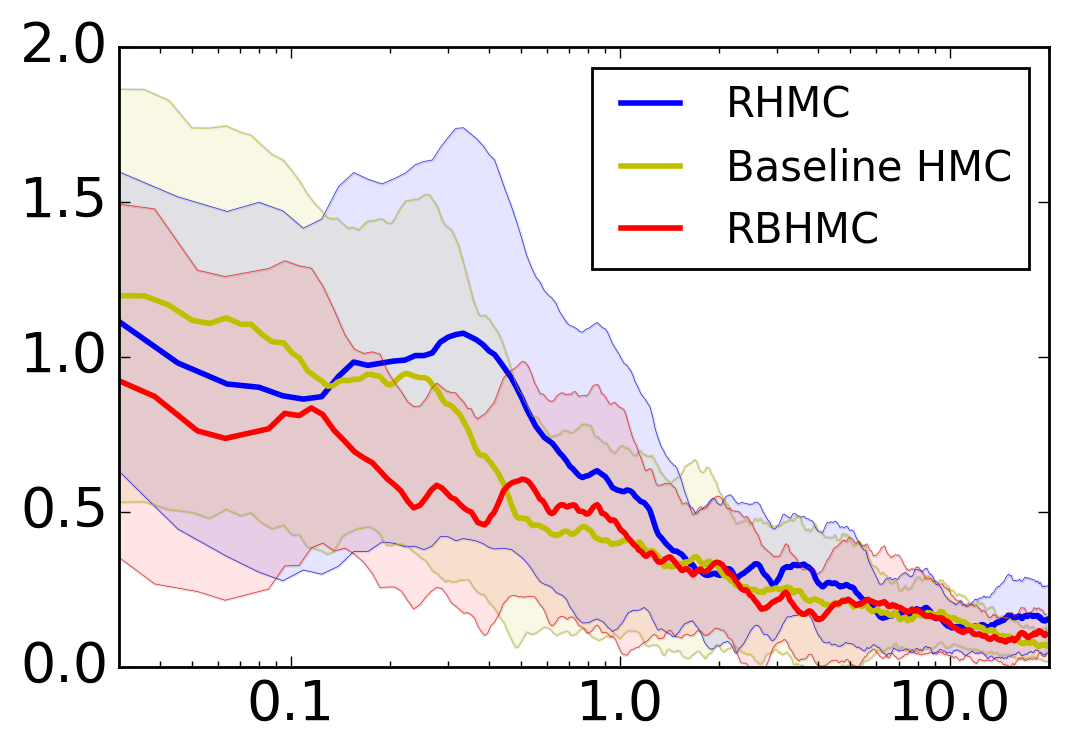}
		\caption{$D = 2$}
	\end{subfigure}
	\begin{subfigure}{0.33\textwidth}
		\centering
		\includegraphics[width = \linewidth]{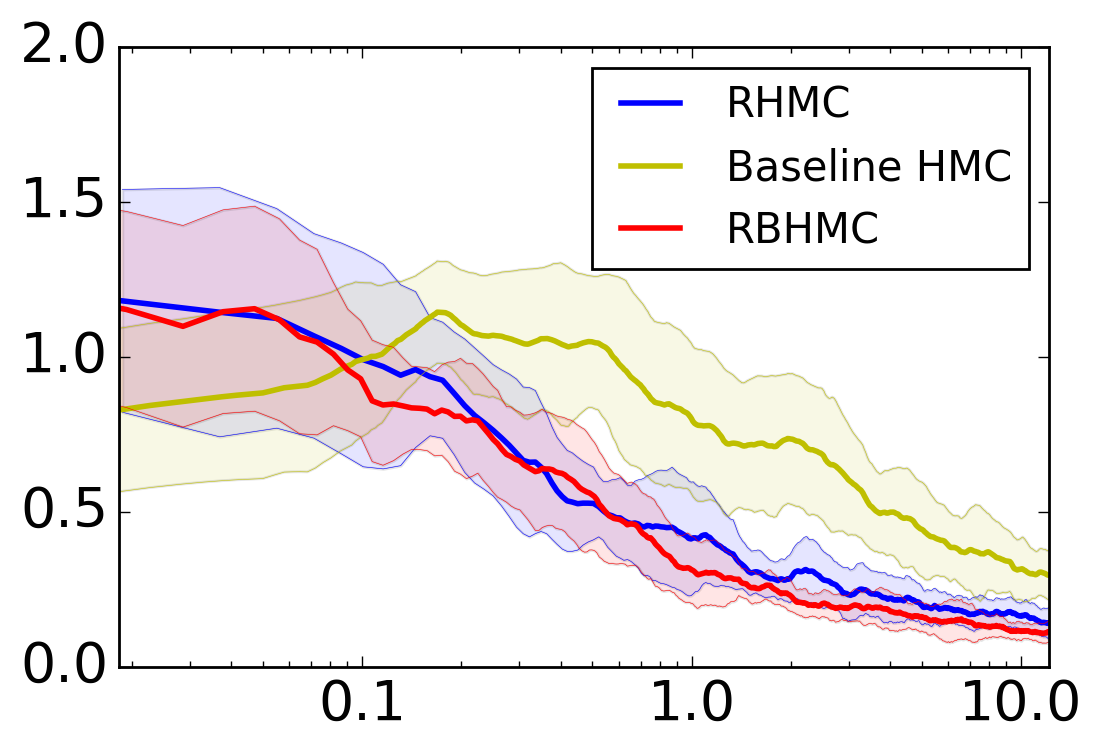}
		\caption{$D = 20$}
	\end{subfigure}
	\begin{subfigure}{0.32\textwidth}
		\centering
		\includegraphics[width = \linewidth]{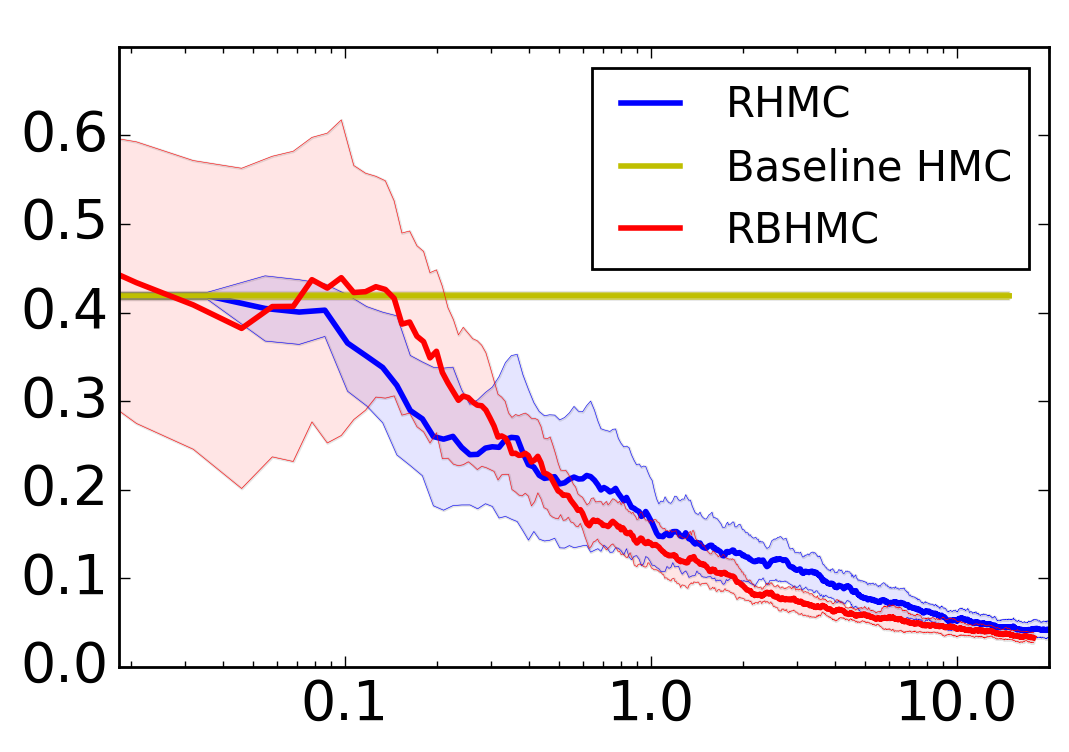}
		\caption{$D = 50$}
	\end{subfigure}
	\caption{Worst mean absolute error (WMAE) versus running time (in seconds) for RBHMC, RHMC and baseline HMC under 3 different dimensionalities. When $D = 50$ the baseline HMC is not able to accept any new samples. For all 3 samplers, $\varepsilon = 0.0167$, $L = 600$, $\mu = 100$.}
	\label{fig:WMAE}
\end{figure}

Under three different dimensionalities $D = 2, 20, 50$, each sampler is run for 10 rounds. For simplicity, we set matrix $\mathbf{A}$ to be diagonal, whose elements are randomly selected from $\{\exp(5), \exp(-5)\}$ with equal probability for each round. All samplers are initialized at the same random location. The results are shown in figure \ref{fig:WMAE}. The horizontal axis is the log-scale running time in seconds, and the vertical axis is the WMAE of all samples drawn up to the current iteration. The thick lines are the average WMAE values over the 10 rounds. The color shaded area represents the $99\%$ confidence region. The results suggest that RBHMC and RHMC perform better in high dimension than baseline HMC, since in higher dimensions the particle has more chance to move out of ROI and baseline HMC always rejects these trajectories. In this particular set of experiments, RBHMC performs slightly better than RHMC under all dimensionalities. We reemphasize that even though the computational cost for both methods are comparable to each other and strongly dependent to the shape of truncation boundaries, RBHMC requires much less effort to implement than RHMC.

\subsection{Bayesian Non-negative matrix factorization}

Many Bayesian models have truncated prior distributions on the latent variables in order to constrain their values within certain desired range. For example, consider the following simplified version of the Bayesian model for non-negative matrix factorization (NMF) \cite{Schmidt2009}
\begin{align}
&P(\mathbf{X} | \mathbf{W, A}, \sigma) = \prod\limits_{i,j} \mathcal{N} (\mathbf{X}_{ij} ; (\mathbf{WA})_{ij}, \sigma^2 ) \\
&P(\mathbf{W}) = \prod\limits_{i,j} \lambda_W \exp (- \lambda_W \mathbf{W}_{ij}) u(\mathbf{W}_{ij}) \qquad P(\mathbf{A}) = \prod\limits_{i,j} \lambda_A \exp (- \lambda_A \mathbf{A}_{ij}) u(\mathbf{A}_{ij})
\end{align}
where $\mathbf{X}$ is an $N$ by $D$ observation matrix. The goal is to factorize $\mathbf{X}$ into the product of an $N$ by $K$ matrix $\mathbf{W}$ and a $K$ by $D$ matrix $\mathbf{A}$. It is required that elements of both $\mathbf{W}$ and $\mathbf{A}$ are non-negative. This constraint is imposed by the exponential prior on $\mathbf{W}$ and $\mathbf{A}$, which causes a discontinuous drop in probability density at 0. 

\begin{figure}
	\centering
	\begin{subfigure}{0.32\textwidth}
		\centering
		\includegraphics[width = 0.7\linewidth]{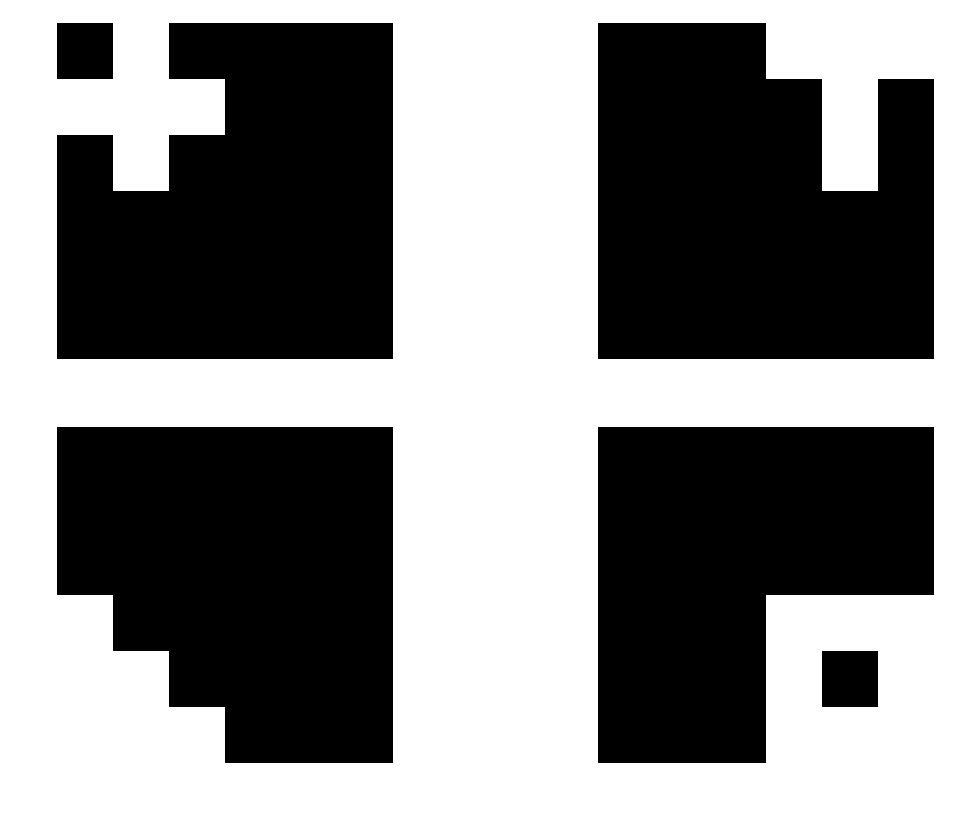}
		\caption{}
	\end{subfigure}
	\begin{subfigure}{0.32\textwidth}
		\centering
		\includegraphics[width = 0.7\linewidth]{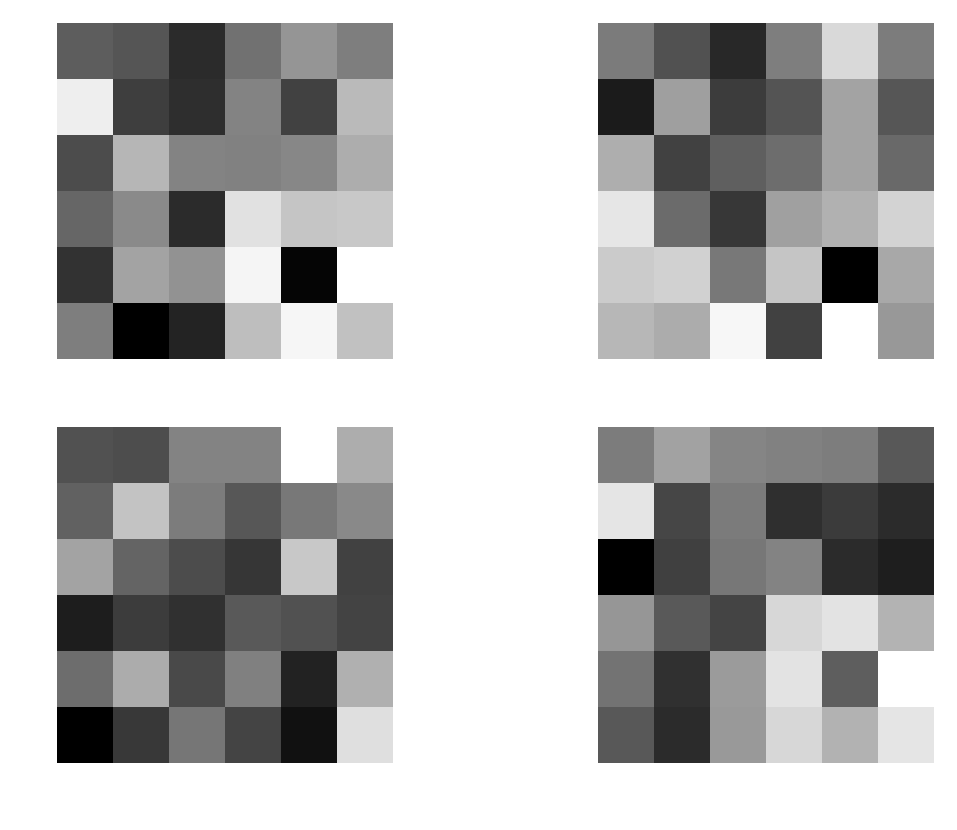}
		\caption{}
	\end{subfigure}
	\begin{subfigure}{0.32\textwidth}
		\centering
		\includegraphics[width = 0.7\linewidth]{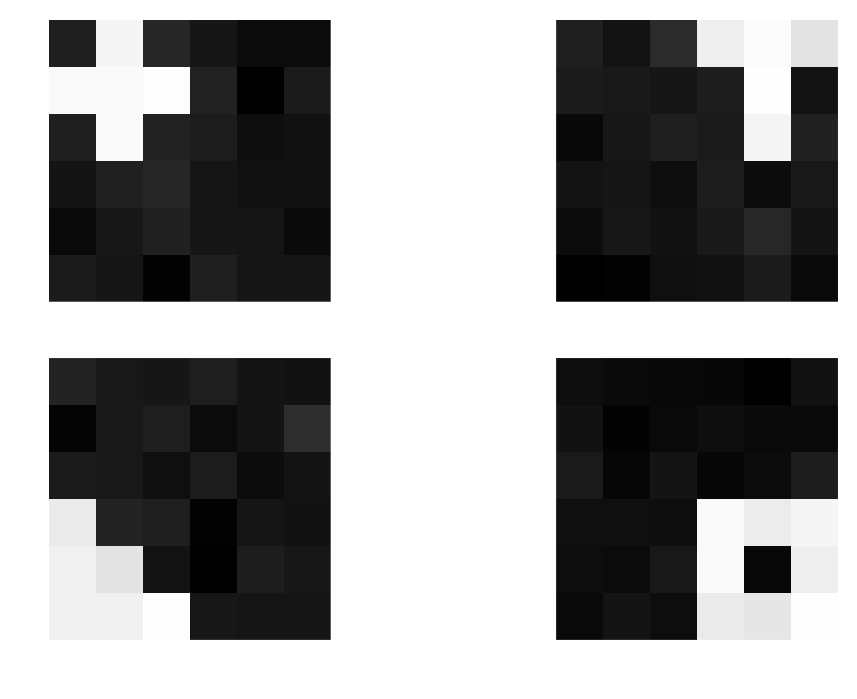}
		\caption{}
	\end{subfigure}
	\caption{(a) Base images from which the observation matrix is generated. Each bright pixel corresponds to value 1 in the matrix element. (b) Examples of generated observation images (c) Sampled base images from RBHMC.}
	\label{fig:GG_block}
\end{figure}

Here we demonstrate the usage of RBHMC to sample from the posterior of $\mathbf{W}$ and $\mathbf{A}$. By replacing the unit step function by a sigmoid factor, the approximate potential function for Bayesian NMF can be written as
\begin{align}
\begin{split}
\tilde{U}(\mathbf{W}, \mathbf{A}) &= \sum\limits_{i,j} \bigg\{ \frac{1}{2 \sigma^2} (\mathbf{X}_{ij} - \sum\limits_k \mathbf{W}_{ik} \mathbf{A}_{kj})^2 + \lambda_W \mathbf{W}_{ij} + \lambda_A \mathbf{A}_{ij} \\
& \quad + \log[1 + \exp (-\mu \mathbf{W}_{ij})] + \log[1 + \exp (-\mu \mathbf{A}_{ij})] \bigg\}.
\end{split}
\end{align}
We run the sampler on a synthetic data set of $N = 1000$ 6 by 6 images generated by 4 base images shown in figure \ref{fig:GG_block}a. Each image is associated with 4 randomly drawn binary numbers $1/0$ which represents the presence/absence of a corresponding base image in the data. The resulting observation matrix $\mathbf{X}$ is the product of an $N$ by 4 binary matrix $\mathbf{W}$ and a 4 by 36 feature matrix $\mathbf{A}$ whose rows are the rearranged pixel values of the base images. A Gaussian noise $\mathcal{N}(0, 0.5)$ is also added to the generated images (figure \ref{fig:GG_block}b). RBHMC simultaneously draws samples of $\mathbf{W}$ and $\mathbf{A}$ from the posterior. Figure \ref{fig:GG_block}c shows the 4 base images reconstructed from one sample of matrix $\mathbf{A}$ drawn by RBHMC when $K = 4$. The result indicates a clean restoration of the original base pattern in the posterior sample. 

\begin{figure}
	\centering
	\includegraphics[width = 0.5\linewidth]{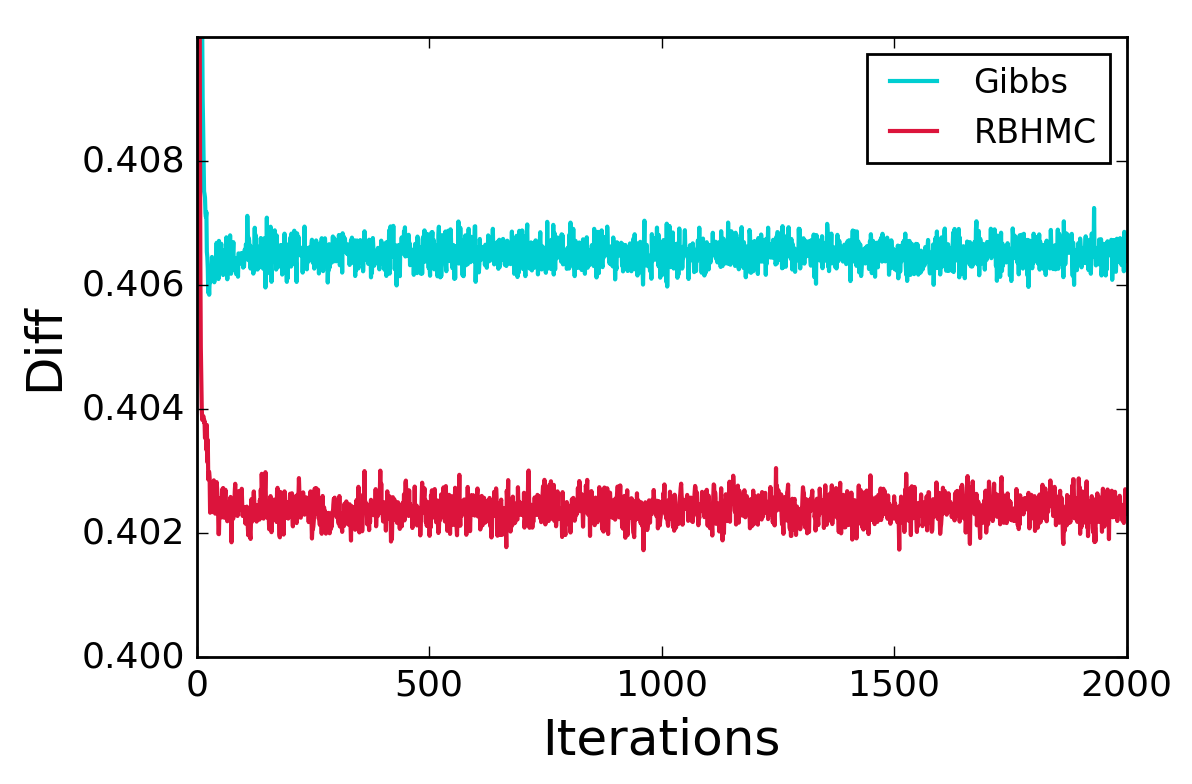}
	\caption{Mean absolute difference versus iteration for RBHMC and Gibbs sampler.}
	\label{fig:Likelihood_plot}
\end{figure}

To further quantify the performance of the sampler, we use the mean absolute difference between the observation matrix $\mathbf{X}$ and the product of the sampled component matrices $\mathbf{W, A}$ as the error measure:
\begin{equation}
\text{Diff}(\mathbf{W, A} ; \mathbf{X}) = \frac{1}{ND} \sum\limits_{i, j} |\mathbf{X}_{ij} - (\mathbf{WA})_{ij}|.
\end{equation}
We compare the performance of RBHMC with Gibbs sampler, which is the baseline inference method introduced in the original proposal of the model \cite{Schmidt2009}. Each sampler is run 10 times for 2000 iterations under $K = 4, \varepsilon = 0.002, L = 200, \mu = 200, \lambda_A = \lambda_W = 1, \sigma = 0.5$. The average mean absolute difference over the 10 rounds at each iteration is shown in figure \ref{fig:Likelihood_plot}. The average difference over all rounds and all samples drawn after the 100th iteration for RBHMC is 0.4023819. The standard deviation across the 10 rounds averaged over the last 1900 iterations is 0.0013394. For Gibbs sampler, the average difference is 0.4065109, and the average standard deviation is 0.0013314. The results show a statistically significant difference in performance between the two methods, which suggest that RBHMC is able to draw samples that better reconstructs the observation matrix than the Gibbs sampler, and therefore serves as a nice alternative inference method for the Bayesian model.

\section{Conclusion}

In this work, we introduce RBHMC as an extension of Hamiltonian Monte Carlo for sampling from truncated distributions. Despite the fact that there exists some numerical restrictions, we have demonstrated through our discussions the following advantages of the new method. First, RBHMC provides an easy-to-implement method for sampling from arbitrarily truncated distributions in high dimensions, with the only requirement being that the density function is smooth over the sample space. Second, RBHMC achieves comparable computational efficiency compared to other existing HMC-based methods on constrained spaces. Third, RBHMC gives rise to a new inference method for Bayesian models with truncated prior distributions. Furthermore, as a potential direction for future research, since conjugacy is not required for HMC, our method has rich potential applications in non-conjugate constrained space Bayesian models.

\newpage
\bibliographystyle{unsrt}
\bibliography{references}
\end{document}